\documentclass[%
reprint,
amsmath,amssymb,
aps,
]{revtex4-2}
\usepackage{graphicx}% Include figure files
\usepackage{dcolumn}% Align table columns on decimal point
\usepackage{bm}% bold math
\usepackage[colorlinks=true,linkcolor=blue,anchorcolor=blue, citecolor=blue,urlcolor=blue]{hyperref}% add hypertext capabilities
\usepackage[mathlines]{lineno}% Enable numbering of text and display math
\usepackage{tikz}
\newcommand*\circled[1]{\tikz[baseline=(char.base)]{\node[shape=circle,draw,inner sep=2pt] (char) {#1};}}
\begin{document}
%======================== head
\title{Deep-CNN based Robotic Multi-Class Under-Canopy Weed Control in Precision Farming}% Force line breaks with \\
% \thanks{A footnote to the article title}%

\author{Yayun Du$^{1}$,Guofeng Zhang$^{1}$, Darren Tsang$^{1}$}
%  \altaffiliation{Physics Department, XYZ University.}%Lines break automatically or can be forced with \\
\author{M. Khalid Jawed$^{1,}$}%
\email{Correspondence should be addressed to khalidjm@seas.ucla.edu}
\affiliation{%
	$^{1}$Department of Mechanical \& Aerospace Engineering, \\ University of California, Los Angeles
}%
%=======================Abstract=====================
\begin{abstract}
Smart weeding systems to perform plant-specific operations can contribute to the sustainability of agriculture and the environment. Despite monumental advances in autonomous robotic technologies for precision weed management in recent years, work on under-canopy weeding in fields is yet to be realized. A prerequisite of such systems is reliable detection and classification of weeds to avoid mistakenly spraying and, thus, damaging the surrounding plants. Real-time multi-class weed identification enables species-specific treatment of weeds and significantly reduces the amount of herbicide use. Here, our first contribution is the first adequately large realistic image dataset \textit{AIWeeds} (one/multiple kinds of weeds in one image), a library of about  10,000 annotated images of flax and the 14 most common weeds in fields and gardens taken from 20 different locations in North Dakota, California, and Central China. Second, we provide a full pipeline from model training with maximum efficiency to deploying the TensorRT-optimized model onto a single board computer. Based on \textit{AIWeeds} and the pipeline, we present a baseline for classification performance using five benchmark CNN models. Among them, MobileNetV2, with both the shortest inference time and lowest memory consumption, is the qualified candidate for real-time applications. Finally, we deploy MobileNetV2 onto our own compact autonomous robot \textit{SAMBot} for real-time weed detection. The 90\% test accuracy realized in previously unseen scenes in flax fields (with a row spacing of 0.2-0.3 m), with crops and weeds, distortion, blur, and shadows, is a milestone towards precision weed control in the real world. We have publicly released the dataset and code to generate the results at \url{https://github.com/StructuresComp/Multi-class-Weed-Classification}.	
\end{abstract}
\maketitle

%\tableofcontents
%========================Introduction====================
\section{Introduction}
Herbicides can have negative side-effects on the ecosystem, biodiversity, and human health~\cite{horrigan2002sustainable}. Conventional weed control methods indiscriminately spray the entire field, including soil, crops, and weeds, with a single herbicide. This strategy is widely applied as it does not require the users to know the spatial distribution or the type of the weeds. However, overuse of chemicals has led to hundreds of herbicide-resistant species of weed across the world~\cite{lebaron1990herbicide}. Reducing the amount of herbicides is a crucial step towards sustainable agriculture. Site-specific weed control can result in 90\% savings in herbicide expenditures~\cite{timmermann2003economic}. 
% In addition to that, real-time weed classification can enable species-specific control strategies, e.g. a robot can adjust the type and amount of herbicide based on the the kind of weed detected. This can further reduce herbicide use. 
Since the worldwide annual sales of pesticides are on the order of a hundred billion dollars~\cite{carvalho2017pesticides}, the economic impact -- in addition to environmental benefits -- of precision weed management is overwhelming.
\begin{figure*}
	\centering
	\includegraphics[width=0.6\textwidth]{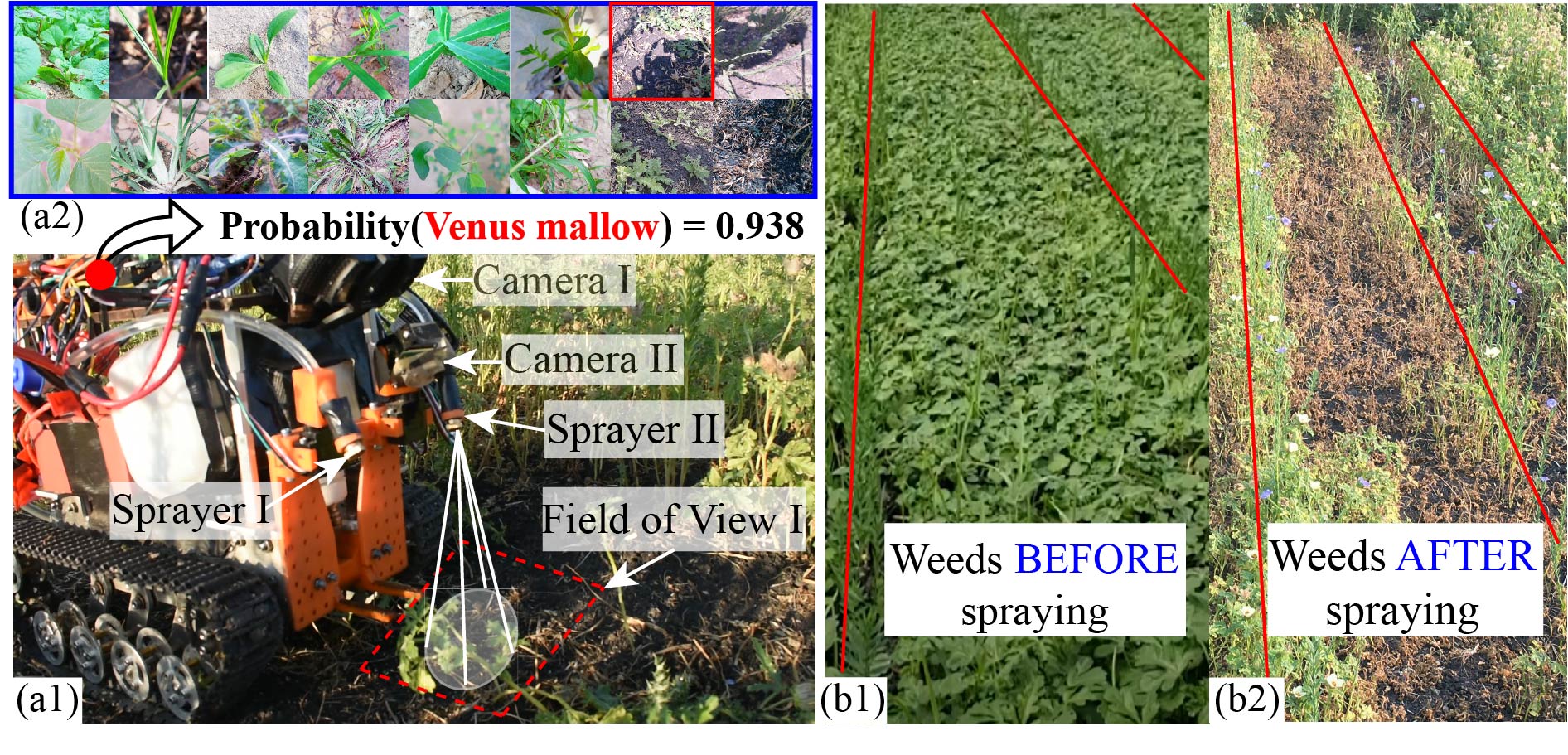}
	\caption{(a1) The cameras on our \textit{SAMBot} robot performing multi-weed classification to recognize the weed(s) in the view of camera I (shown as the dashed red rectangle) in the flax fields by exploiting the light-weight MobileNetV2 CNN model and executing spraying (continuous white lines); (a2) The predicted weed -- Venus mallow (in the small red rectangle) in the view of camera I out of the 16-class \textit{AIWeeds} (in the blue rectangle) by onboard Jetson Nano (denoted as the filled red dot in (a1)); Snapshots showing the statuses of weeds (b1) before and (b2) after robotic herbicide spraying. Continuous lines in both (b1) and (b2) represent the flax croplines. Note that weed classification is for point spray but not for this dense weed scenario, so (b1) and (b2) are here to validate our robotic spraying system.
	}
	\label{fig:overview}
\end{figure*}
% \sidecaptionvpos{figure}
% \begin{SCfigure*}[25][t]
% \centering
% \includegraphics[width=0.6\textwidth]{figures/Figure1.jpg}
% \captionsetup{width=1.0\textwidth}
% \caption{(a1) The cameras on our \textit{SAMBot} robot performing multi-weed classification to recognize the weed(s) in the view of camera I (shown as the dashed red rectangle) in the flaxseeds fields by exploiting the light-weight MobileNetV2 CNN model and executing spraying (continuous white lines); (a2) The predicted weed -- Venus mallow (in the small red rectangle) in the view of camera I out of the 16-class \textit{AIWeeds} (in the blue rectangle) by onboard Jetson Nano (denoted as the filled red dot in (a1)); Snapshots showing the statuses of weeds (b1) before and (b2) after robotic herbicide spraying. Dashed red lines in both (b1) and (b2) represent the flax croplines. Note that weed classification is for point spray but not for this dense weed scenario, so (b1) and (b2) are here to validate our robotic spraying system.}
% \label{fig:overview}
% \end{SCfigure*}
The past decade has seen revolutionary advances in mobile robotic platforms for precision agriculture and several commercial entities (Blue River Technology, ecoRobotix, Hitch Robotics, EarthSense, and Naio Technologies) have commercialized various robotic vehicles. Real-time weed control is the primary objective of several of these vehicles, e.g. ``See and Spray Machines" of Blue River Technology and TerraSentia of EarthSense. In the context of these ongoing activities in commercial as well as academic~\cite{lottes2018joint} sectors, real-time weed detection and classification is a truly enabling technology for robotics. However, three critical challenges -- (1) data, (2) training, and (3) real-world deployment -- have to be overcome for fully autonomous weed management. While many weed libraries have been released~\cite{dos2017weed,dyrmann2016plant, olsen2019deepweeds, lee2015deep, chebrolu2017agricultural}, what is lacking is a large enough image dataset of realistic fields, including crops and weeds, distortion, shadows, and motion blur. The onboard computing power of agricultural robots can often be limited, especially for row crops (e.g. flaxseed and canola) where the inter-row spacing is small as 0.2-0.3 m. This spacing limits the size of the robot.
Existing platforms except for robots from ecoRobotix are suitable only for fields with wide row spacing ($\geq0.4$ m) as they use GPS (limited accuracy with crop blocking and size-incompatible with small robots) for navigation and guidance.  Moreover, the robot should be able to travel under-canopy once the crop canopy has been established for truly precise weed management. This further restricts the size of the robot as well as the computing capability. Small vehicles to travel under-canopy have been introduced by EarthSense. However, owing to the challenges above, autonomous weed management with small vehicles is yet to be achieved.

% Smart, perception-controlled weeding systems can contribute to herbicide reduction by handling the weeds on a plant-specific level and selectively spraying different species since they show variable sensitivities to different herbicides. Precise weed detection and classification, however, poses challenges from both the hardware and algorithm aspects. Giant platforms such as BoniRob~\cite{lottes2018joint} and Naio weeding robot~\cite{naio}, equipped with powerful computing unit and high-resolution cameras, are costly and not adaptable to most fields with 20-30 cm spacing between rows of crops. By contrast, UAVs are much smaller, cheaper, and can cover larger areas in a comparatively shorter period of time than ground vehicles, but they can only fly/hover at a relatively high position for a limited amount of time. Both UAVs and large robots have limited applicability once the crop canopy has been established. Pixel-wise segmentation algorithms require cumbersome manual data labeling and are computationally expensive. Therefore, these algorithms are mostly used with pre-recorded videos and are not realistically deployable in real-time systems.
To address these issues, we employ our own low-cost (around \$500) compact robot, \textit{SAMBot}, as shown in Fig. \ref{fig:overview}(a1), to which two cameras are attached $0.2-0.4$m above the ground. The customized three-degree-of-freedom gimbal enables the camera to scan an angle between $0^{\circ}$ to $150^{\circ}$.
% The low distance from the ground reduces the number of instances entering the view of the camera.
% Details of hardware implementation will be given in Section \ref{sec:experiment}. Experiments in Fargo, North Dakota during the summer of 2019 showed that the robot can navigate in flaxseed fields with one-foot spacing between rows and its spraying system works perfectly. Being low-cost, compact, and effective, targeted weed spraying with our robot not only benefits the ecological health of the farmland but can also automate weed removal in household lawns.
We employ a mobile Convolutional Neural Network (CNN) for weed classification. We chose this CNN because it is the next-generation of on-device computer vision networks, can predict much faster than other networks, and maintain competitive performance. The network utilizes a modified MobileNetV2~\cite{sandler2018mobilenetv2} architecture (i.e. dropping last two fully-connected layers and adding one global average pooling layer) and its inference is optimized by NVIDIA TensorRT. Using the same pipeline, we also train and run another 4 models with acceptable training speed and accuracy. The contributions of this system paper are as follows:
\begin{itemize}
	
	\item We release a realistic dataset -- \textit{AIWeeds} -- containing 10,000 labeled images with multiple weeds/crops in view, including flax and 14 most common weeds in North Dakota, the middle of China, and California~\cite{weedsData}. The images are taken at several stages of weed from 20 different sites in the fields and gardens with variations in illumination, shadows, weather conditions, view perspective and distance, and plant growth stages. 
	% This library can be used not only for classification and object detection but also for other plant studies.
	
	\item We demonstrate a multi-class weed classification pipeline for model deployment on GPU boards, even low-end boards such as Jetson Nano/Xavier Nx. Using this pipeline, we study weed classification using 5 modified models: CNN-MobileNetV2 and 4 other more complicated models and compare their performance through various metrics. Researchers can use this study to evaluate their unique situational needs, such as training time and accuracy, and select the appropriate model for their needs.
	
	\item  We run CNN-MobileNetV2, the only qualified candidate for onboard under-canopy weed classification, on \textit{SAMBot} (with Jetson Nano) in flax fields and achieve satisfactory experimental results, i.e. accurately detect and spray weeds. This is a milestone for real-time autonomous weeding in realistic fields.
	
	% 	\item The $F_1$-score of our modified MobileNetV2 structure reaches 0.9 when training and testing the images from our dataset. Then, we run the CNN on our \textit{SAMBot} robot platform in real time and achieve satisfactory experimental results. Our robot can potentially perform more precise agricultural duties by not only spatially surveying key indicators of crop health and soil moisture but also applying fertilizers to replace labor-intensive manual procedures.
\end{itemize}

The remainder of this study is arranged as follows. Section \ref{sec:relatedWork} presents the state-of-the-art on weed classification. Section \ref{sec:dataCollection} describes how the dataset is built while Section \ref{sec:modeling} gives details of our multi-class weed classification pipeline. We then show our experimental results of models trained on our dataset and deployed onto the real-time robot system. Finally, the conclusion is drawn in Section \ref{sec:conclusion}.

\section{RELATED WORK}
\label{sec:relatedWork}

\subsection{Vision-Based Weed Control \& Datasets}
\label{subsec:vision}

Image-based~\cite{lee2017deep, bakhshipour2018evaluation, dos2017weed}, spectrum-based~\cite{shirzadifar2018weed, li2017design} and spectral image-based~\cite{louargant2018unsupervised,lin2017detection} methods have been successfully applied to identify weeds from both ground and aerial photography. Spectrum and spectral image-based approaches are ideally suited for highly controlled site-specific environments where spectrometers can be tailored for consistent acquisition and detection. Nonetheless, it is challenging to incorporate them in harsh field environment and deploy them onto compact vehicles. Vision-based methods, on the other hand, benefit from cheaper and simpler image acquisition under varying illumination conditions.
% We therefore concentrate on applying vision-based techniques for weed classification to our low-cost miniaturized compact robot that will be introduced in Section \ref{subsec:experimentSetup}. 
Weed datasets in realistic fields are indispensable for vision-based weed control, but the existing ones~\cite{ dos2017weed,dyrmann2016plant, olsen2019deepweeds, lee2015deep} only embrace ideal weed images without background, none of which is apt for real field applications. The most realistic one~\cite{chebrolu2017agricultural} consists of static clear (without motion blur and shadows) images shot from a straight downwards view. Nonetheless, distortion, shadows, and motion blur that always appear during applications increasing technical difficulty are not reflected in~\cite{chebrolu2017agricultural}.

\subsection{Multi-class Classification Using Deep Neural Network}
\label{subsec:CNN}

Classical vision-based weed classification methods rely on different features of crop plants and weeds, such as color, leaf shape~\cite{beghin2010shape} and size, vein patterns, and so forth~\cite{li2019review}. However, in complex natural scenarios with high weed densities where weeds and crop plants overlap and occlude, they cannot perform the task correctly and robustly. In addition, further investigation is needed as to whether they are applicable to actual field conditions. This problem is addressed by recent deep learning models, such as Convolutional Neural Networks (CNNs). CNNs~\cite{olsen2019deepweeds,lottes2016effective, sa2017weednet,milioto2018real, fawakherji2019crop} take advantage of a deep hierarchical structure to extract global features of the image and context information (background such as soil), which significantly reduces the error rate of image recognition than the classical algorithms mentioned above. 
For early-stage wide weed control (around stem elongation but before booting), many current methods for weed detection and classification focus on segmenting images with both RGB and excess green/red or near-infrared~\cite{lottes2018joint,lottes2016effective,lottes2018fully} recordings. However, they are powerless in fields with canopies developed and visual occlusion. Performing early-stage weed control will probably destroy or affect the growth of crops and thus the crop yield. As such, under-canopy weed control is a must. ~\cite{lottes2016effective} and ~\cite{dyrmann2016plant} tried semantic segmentation of weeds and crops. Still, segmentation methods~\cite{lottes2016effective, dyrmann2016plant, lottes2016effective,lottes2018fully} require lots of effort and time because of hand-crafted labeling and are challenging to be run on an embedded computer. As a result, there is a timely need to do under-canopy multi-class weed identification in real fields with CNNs.

\begin{figure*}
	\centering
	\includegraphics[width=.8\textwidth]{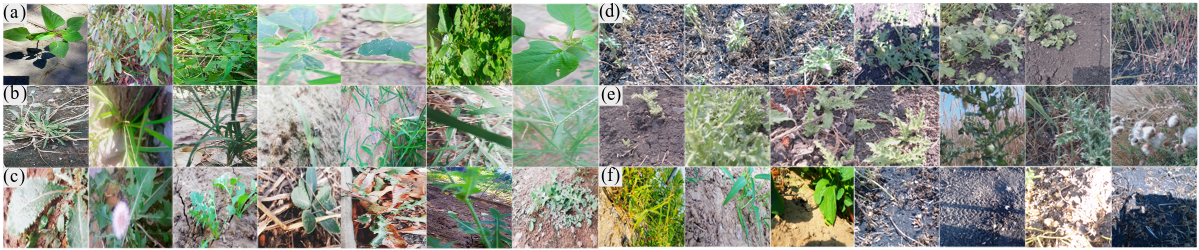}
	\caption{Sample images from some classes of the \textit{AIWeeds} dataset to show the variation: (a) Amaranthus spinosus (AS.), (b) Brachypodium sylvaticum (BS.), (c) Cirsium arvense (CA.), (d) Venus mallow (VM.), (e) Canada thistle (CT.), (f) Negatives (NG.).
	}
	\label{fig:datasetSample}
\end{figure*}

\subsection{Under-canopy Weeding with Low-cost Robots}
%~\cite{deepweeds, 3/18 in real-time weed-crop classification paper}
As mentioned above, CNN models have been widely explored by researchers and run on expensive, powerful PCs~\cite{lottes2016effective, dyrmann2016plant, lottes2018fully, olsen2019deepweeds}. However, none of the prior works deployed their model onto a flexible mobile robot that works for under-canopy weeding. Powerful PCs are usually not an option for small economical platforms.
% It is also not available to our robot because we aimed to build a low-cost, functional weed-control robot. Our previous experience with the powerful and relative expensive Intel NUC8i7BEH (12-19 VDC, 28W) informed us that even under light workload, the operational time on a single charge is around 30 minutes with 14.4V LiPo batteries on board. 
Our prior work verified that our low-cost robot platform \textit{SAMBot} worked autonomously in real fields with narrow row spacing, e.g., flax and canola fields. On \textit{SAMBot}, NVIDIA Jetson Nano is used as a cost effective (\$99) solution that possesses the performance and capabilities to run modern artificial intelligence workloads. 

In summary, to the best of our knowledge, this work is the first trial applying lightweight multi-class models to agricultural robots for under-canopy weed control and testing the performance in real flax fields. 

%
%The paper is structured as follows: we review related work in Sec. II, followed by Sec. III in which we describe the theories and algorithms applied in our approach. Experiments and results are discussed in Sec. IV. Finally, Sec. V concludes the paper.
\begin{table*}[t!]
	\centering
	\caption{The weed species collected in our dataset and their corresponding quantity.}
	\begin{tabular}{|c|c|c|c|c|c|c|c|c|c|c|c|c|c|c|c|c|}
		\hline
		Weed         & AS. & BS. & CT. & CA. & CD. & D. & Flax & L. & Neg. & N. & PM. & SF. & SA. & VM. & VP. \\ \hline
		\# of Images & 659 & 655 & 560 & 990   & 631  & 428  & 625 & 549   & 1474  & 649  & 526   & 566  & 565   & 704 & 559   \\ \hline
	\end{tabular}
	\label{table:weedAndNumber}
\end{table*}
\section{Dataset collection}
\label{sec:dataCollection}

Our first goal is to create a variable and realistic dataset that allows us to step towards the further objective,i.e. to realize and enhance the baseline accuracy of the off-the-shelf CNNs and make it easy to be trained and deployed to facilitate wider use of the dataset. 
Finally, the deployed model enables \textit{SAMBot} to detect, identify, and precisely spray weeds during field marching, even when the visual appearance of the plants and background has changed. In this section, we provide details of how we collected images to reflect varying scenes and target variability in realistic fields. 
% The quality evaluation of our dataset, \textit{AIWeeds}, will be reported in Section \ref{subsec:experimentSetup} reflected by the performance of models. 

\begin{figure*}[t!]
	\centering
	\includegraphics[width=.9\textwidth]{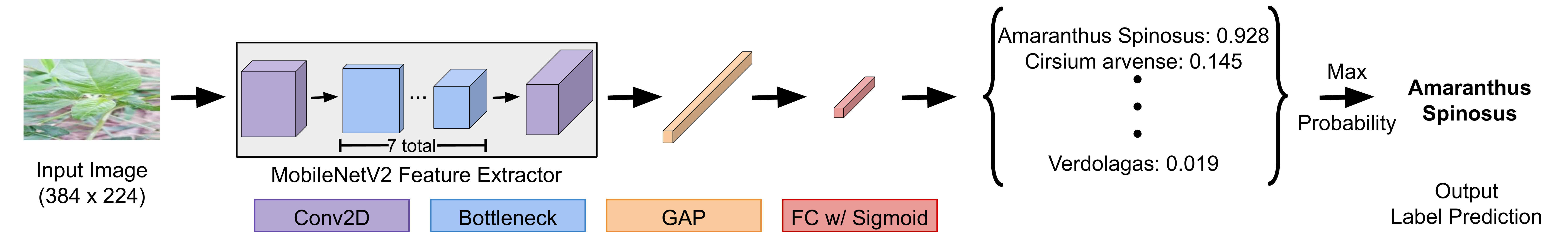}
	\caption{Detailed architecture of learning model for multi-weed classification with modifications on MobileNetV2.
	}
	\label{fig:mobilenetv2}
\end{figure*}

% The performance of any machine learning model, ranging from linear regression to CNNs, is bound by the dataset it is learning from. 
Unlike the aforementioned libraries~\cite{dos2017weed,dyrmann2016plant, olsen2019deepweeds, lee2015deep, chebrolu2017agricultural}, we build a dataset named \textit{AIWeeds} containing flax and 14 most common weeds in the fields in North Dakota, California, and central China. % For the image acquisition, we used Samsung's flagship ``flip phones'' with a resolution of 1080$\times$1920 and 1280$\times$720 pixels.
The image resolution is 1920$\times$1080 or 1280$\times$720 pixels.
Table \ref{table:weedAndNumber} shows the weed species and their corresponding quantity in \textit{AIWeeds}. Full and corresponding abbreviated names of the weeds are: Amaranthus spinosus (AS.), Brachypodium sylvaticum (BS.), Cirsium arvense (CA.), Cynodon dactylon (CD.), Dandelion (D.), Lambsquarters (L.), Nutsedge (N.), Plantago Major (PM.), Setaria faberi (SF.), Sonchus arvensis(SA.), Verdolagas Pirslane (VP.), Venus mallow (VM.), Canada thistle (CT.), Flax, and Negatives (Neg.).
A total of more than 10,000 images were taken under different sunlight (from 7 a.m. to 6 p.m.), weather conditions (super bright, sunny, cloudy, and rainy), growth stages (from sprouting to ripening), and varying health conditions (under drought, plant disease, and insect pest infection). On average, 600 images of each target species were taken from at least three different locations. Rotation and scale of the target weed species in the images also vary as they are photographed \textit{in situ} with unknown orientation. Fig.~\ref{fig:datasetSample} displays some image samples from our dataset and gives a sense of the variations. These variations in conditions were deliberate in order to significantly increase the generality of \textit{AIWeeds}. They can, however, influence the foliage color, strength of features, and other noticeable anomalies. This includes the intra-species variation of the data. As shown in Fig. \ref{fig:datasetSample}, another variability in our dataset arises from the complex and dynamic target backgrounds. Although our dataset is large, over-fitting might still appear, the main concern of deep neural networks. Later in Section \ref{sec:modeling}, we will describe the image augmentation skills implemented to avoid over-fitting. We also kept this in mind while constructing \textit{AIWeeds}. We rotated the camera and rendered motion blur during the shot. These confounding factors of heterogeneity in \textit{AIWeeds} will jointly lead to deeper and more complex models to attain acceptable performance.

In addition, locations subject to dense weed infestations are also populated by other native plants. Since we are unable to process a dataset including all plants, all other non-target species in view must be labeled as \textit{negative} samples, along with all non-target background images. Unfortunately, this introduces a highly variable class in the dataset that will be difficult to classify consistently. In order to prevent over-fitting, increase the accuracy and robustness of weed classification models despite the disturbance of non-target plants, we include the \textit{negative} class in \textit{AIWeeds} (abbreviated as Neg.) in Table \ref{table:weedAndNumber}. 
% by ensuring the identification of targeted weeds from their surrounding backgrounds

% In summary, we concentrated on two primary aspects when establishing an expertly labeled dataset with the requisite variability and generality. First, collect on average 600 images of each target species from at least three different locations. 
% Second, create a \textit{negative} class to contain the non-target background from each location. The first aspect is a necessity for deep and high-complexity because CNN models require large labeled datasets when training. The second aspect aims at preventing the over-fitting of developed models with scene-level features by ensuring the identification of targeted weeds from their surrounding backgrounds. 
% Every image is cautiously identified as to whether it contains one/multiple targeted weeds before being labeled. 
The strictness of the collection process will ensure the accuracy and robustness of all classification learning models. Fig. \ref{fig:datasetSample} displays a subset of \textit{AIWeeds} to demonstrate its variety and generality within classes, from which the complexity of the learning problem is evident.  

\section{Multi-class Weed Classification Pipeline}
\label{sec:modeling}
% \subsection{Model Training}
% \label{subsec:training}
In this section, we illustrate the pipeline for a multi-class weed classification scheme. To implement and realize our models, we utilized the popular machine learning framework, Tensorflow, and high-level API, Keras. This allowed us to try out various models, such as VGG19, NasenetMobile, ResNet50, InceptionV3, Xception, DenseNet, and MobileNetV2. After testing different models, we concluded that MobileNetV2 provided the best combination of low memory usage and computational time but maintained a respectable level of accuracy. It is, therefore, deployable onto Jetson Nano with limited onboard computation resources.
% We will be focusing on MobileNetV2 in the following explanation. 
% MobileNetV3~\cite{howard2019searching} and EfficientNet~\cite{tan2019efficientnet} are two newer and faster models than MobileNetV2 but MobileNetV3 is based on Tensorflow 2.0, which is incompatible with the version of TensorRT that comes with the Jetson Nano base operating system. EfficientNet requires the input images to be square in size, which is not applicable to our dataset. Fortunately, MobileNetV2 already achieves adequate performance while running smoothly on the Jetson Nano board, as seen later in Section \ref{sec:experiment}. In addition,

% FCN~\cite{fawakherji2019crop, lottes2018fully} and semantic segmentation models were not considered as they ask for within-image time-consuming per-pixel labels and not yet available for \textit{AIWeeds}.

All models were pretrained to recognize 1,000 object classes in ImageNet, and we slightly modified their architectures to classify 16 classes (14 types of weeds, flax, and negatives) in \textit{AIWeeds}. Modifications will be illustrated on MobileNetV2 because they are identical for other models. As shown in Fig. \ref{fig:mobilenetv2}, the last fully connected layer consisting of 1,000 neurons of ImageNet-trained MobileNetV2 is replaced by a 16-neuron fully connected layer. We used MobileNetV2 as a feature extractor and added two layers at the end: a global average pooling (GAP) layer and a fully connected layer that used sigmoid as the activation function. Sigmoid is chosen because the probability of every class presenting in the same view (image) in nature is the same. It allows the output layer to identify the likelihood of an image belonging to each class. The weed with the highest sigmoid-activated neuron probability is thought to appear in the input image.

% When one of the weeds has a high probability to be predicted as the output, others should have a smaller probability since their probability sum is one. The final output will be the weed with the highest softmax-activated probability from the fully connected layer.

The preprocessing preparation for learning includes image flipping, resizing, and augmentation. First, all images for  training, validation, and testing were rotated and resized to $384 \times 224$, the size closest to the default $224 \times 224$ in ImageNet while matching the normal output of the robot's camera and keeping the ratio of our photographs taken to prevent excessive distortion. Next, image augmentation was performed by rotating every image arbitrarily in the range of [$-360^\circ, 360^\circ$] and then scaling it in the range of [0.5, 1] both horizontally and vertically. 
After that, each color channel and pixel intensity were both randomly shifted between -25 pixels to 25 pixels to account for the illumination variance. We also randomly scaled pixel intensity within [0.75, 1.25] range and did random perspective transformations on each image to stimulate a wide range of viewing distances and angles. In summary, our image augmentation implementations accounted for variations in rotation, scale, illumination, color, and perspective. Otherwise, the deep neural network models mentioned above, with trainable weights in the order of millions, would dramatically over-fit the images by memorizing the training subsets. 

Then, we split the labeled images in \textit{AIWeeds} into training, validation, and testing sets; these sets contained 60-20-20 percent respectively for $k$-fold cross validation with $k=5$. Stratified random partitioning was executed to ensure even distribution of different weed classes within each subset. 
% A random seed was generated to control the random splits of each fold such that the individual split could be reproduced.
A random split of 60\% formed the training dataset, while 20\% constituted the validation dataset to monitor the training process and minimize over-fitting. The remaining 20\% were reserved for testing and never allowed to join any training procedure. Normally, training the models from scratch with our custom dataset cannot guarantee acceptable performance even after a long training time on a computationally powerful platform. As a result, each model was loaded  with its corresponding pre-trained weights on ImageNet as initial weights before training through Keras. The weights of the fully-connected layer were initialized by uniform distribution.

Finally, we fine-tuned the layers using our custom built dataset. The standard binary cross-entropy loss function and Adam optimizer were used to train all models. Batches of 32 images were produced for training, which would be aborted if the validation loss did not decrease after 32 epochs. Here, the validation loss refers to the classification error calculated on the validation subset of images. The training was restarted after an abortion by loading the continuously saved model with the smallest running validation loss. After exploration, the initial learning rate was set as 0.0001  and was then successively halved every time the validation loss did not decrease after 16 epochs. The learning rate would be reduced to $0.5 \times 10^{-4}$ when the training restarted after an abortion. The validation and testing results of all models will be given in Section \ref{subsec:quantitative}. 

% \subsection{Model Deployment on Embedded Boards}
% \label{subsec:deployment}
Experiments demonstrated that only MobileNetV2 could be deployed and run on hardware-limited \textit{SAMBot} (with Jetson Nano) with/without structure optimization by TensorRT, while other complicated models are not deployable even with TensorRT speedup. We used TensorRT to optimize the inference time that delivers low latency, memory usage, and high throughput.
The four key operation steps related to TensorRT include creating frozen graphs for trained models, converting frozen graph to the TensorRT engine, running TensorRT engine, and benchmarking all models. Details are given in our open-source repository.
% First, we converted the saved hdf5 file of the trained model to a regular pb file, which was then frozen into a graph. When we converted the frozen graph to TensorRT engine, we set the precision mode as floating point 16 to replace the original floating point 32 mode to reduce the inference time, and ensured that the inference graph only passed one image. Next, during the execution of the TensorRT graph, a sequence of operations was carried out accordingly. We read the graph, set GPU usage to be expandable, read all images from dataframe of a csv, preprocessed each image by converting it into an array to form batches easily later, and generated the image prediction by linking it to the input tensor name and making predictions via tf.session with the name of the output tensor. In the end, we recorded the time needed for preprocessing and inferencing for each image and output the respective confusion matrix and classification report, which will be covered in Section \ref{subsec:quantitative}.

\begin{figure}[h!]
	\centering
	\includegraphics[width=\columnwidth]{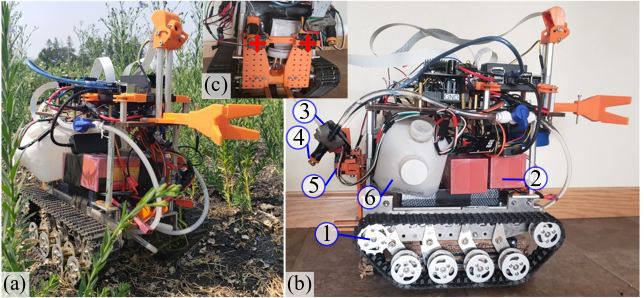}
	\caption{(a) \textit{SAMBot} working in flax fields;(b) The side-view of the hardware layout of \textit{SAMBot}; (c) The back-view of the adjustable "see and spray" system on \textit{SAMBot}.
	}
	\label{fig:robot}
\end{figure} 
\section{Experiments \& Results}
\label{sec:experiment}

In this section, we illustrate our experimental setup,  followed by the quantitative assessment of five 16-class classification approaches. Then, we deploy MobileNetV2, onto our embedded board, Jetson Nano (the red dot in Fig. \ref{fig:overview}(a1)). Finally, we run our robot, in flax fields in North Dakota, with real-time video streaming. This is to verify the capabilities of the deployed model, i.e. whether the robot can successfully detect multiple types of weeds in flax fields with a medium weed density and spray the corresponding herbicide. The experimental results and supplementary video validate the practicality of MobileNetV2 on \textit{SAMBot}.

\subsection{Experimental Setup}
\label{subsec:experimentSetup}
All experiments were conducted on our miniaturized, low-cost, functional agricultural robot -- \textit{SAMBot} -- in Fig.~\ref{fig:overview}(a1) and Fig.~\ref{fig:robot}. It is developed and tested for weed control in flax (as shown in Fig.~\ref{fig:robot}(a1)) and canola fields of North Dakota, the leading producer with 91\% of the U.S. flax production and 85\% of canola production~\cite{USDANASS}. The robot is generally applicable to row crops. The robot has a powerful drive train \circled{1} (the rated torque of motors is 4.81N$\cdot$m). It successfully passed all the bumps/dents (the maximum height/depth of which is the same as the height of chassis) and finished the full exploration of the fields during field tests during the Summer of 2021 in Fargo, North Dakota. It continuously worked in the fields more than 14 hours with 12-cell 18400mAh onboard LiPo batteries \circled{2}. Referring to Fig.~\ref{fig:overview}(a1) and Fig.~\ref{fig:robot}(b), two cameras \circled{3} are mounted at the back of the robot, about 20-40 cm above the ground. The images in \textit{AIWeeds} were therefore taken from the robot's perspective. One camera and one pressure sprayer \circled{4} are rigidly coupled and actuated together by a gimbal \circled{5} to realize a yaw angle of $0^{\circ}-150^{\circ}$. Each sprayer is connected to a herbicide tank \circled{6}. Not only the orientation of the gimbal assemblies can be changed by servos, but also their positions are adjustable left and right, up and down on the 3D-printed pegboard based on the growth stage of crops/weeds and row spacing, as indicated in the red arrows in Fig.~\ref{fig:robot}(c).
\subsection{Quantitative Results of Workstation Training}
\label{subsec:quantitative}
We use the $F_1$ score 
% that captures both precision and recall performance as follows 
for quantitative evaluation:

\begin{equation} \label{eq:performanceEvaluation}
% \begin{split}
F_1(s) = 2 \cdot \frac{\textrm{precision}_{s} \cdot \textrm{recall}_{s}}{\textrm{precision}_{s} + \textrm{recall}_{s}},  
% \textrm{precision}_s&= \frac{TP_{s}}{TP_{s} + FP_{s}},  \\
% \textrm{recall}_{s}&= \frac{TP_{s}}{TP_{s} + FN_{s}}. 
% \end{split}
\end{equation}
where $\textrm{precision}_{s}$, $\textrm{recall}_{s}$ 
% $TP_{s}$, $FP_{s}$, and $FN_{s}$ 
are the precision, recall
% number of true positives, false positives, and false negatives 
for class $s$, respectively. All models presented in this section are trained and tested with the dataset in Table \ref{table:weedAndNumber}.

\begin{figure}[b!]
	\centering
	\includegraphics[width=.9\columnwidth]{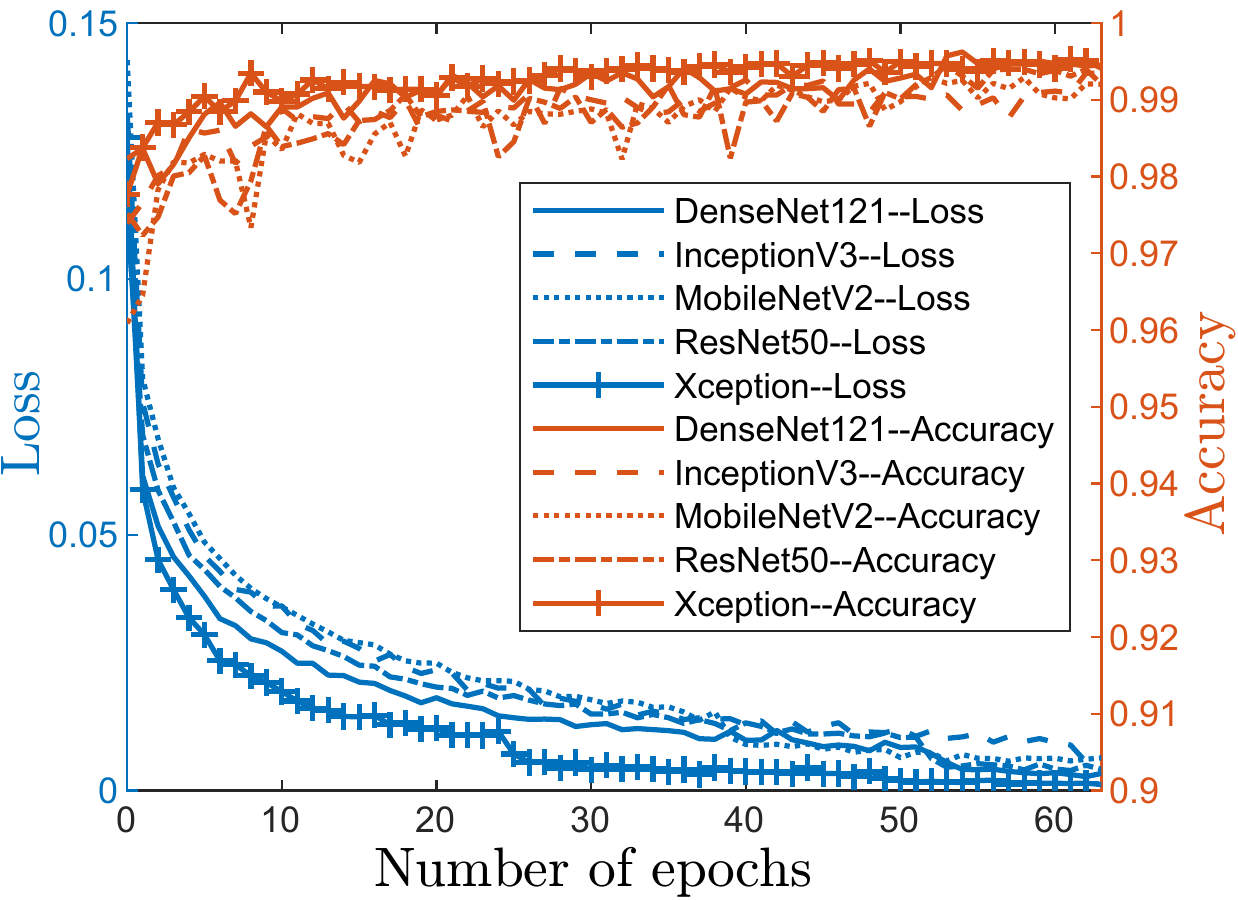}
	\caption{\textbf{Training loss} and \textbf{average validation class accuracy} of different models. The maximum number of epochs is set to 64 for plotting but in reality, MobileNetV2  was trained for 128 epochs.}
	\label{fig:accuracyLoss}
\end{figure}

Fig. \ref{fig:accuracyLoss} shows the training loss and average validation class accuracy of models over 64 epochs on our \textit{AIWeeds} dataset. 
% As the running time of each epoch of different models varies a lot on different hardware platforms, we only provide the performance of each model versus the number of epochs. 
On our GTX 1080 Ti platform, each epoch of DenseNet121, InceptionV3, MobileNetV2, ResNet50, and Xception took 978s, 437s, 274s, 672s, and 672s on average, respectively. \textbf{Training loss} and \textbf{average validation class accuracy} of different models are plotted in Fig. \ref{fig:accuracyLoss} to eliminate over-fitting during training. MobileNetV2 offers an accuracy of 94.50\%, while being the fastest to finish training, far ahead of other models. DenseNet121, on the other hand, gives the best accuracy, 96.77\%, and the smallest loss, 0.0027, after being trained for 64 epochs. We continued to train MobileNetV2 for 128 epochs, which doubled the time needed for 64 epochs, and give a second-best accuracy of 96.15\%.

% \begin{table*}[h!]
% \centering
% \caption{The inference time and resulting accuracy of models after being deployed onto Jetson Nano.}
% \begin{tabular}{|c|c|c|c|c|c|c|c|c|c|c|c|c|}
% \hline
% CNN Models         & DenseNet121 & InceptionV3 & MobileNetV2 & ResNet50 & Xception  \\ \hline
% Inference Time w/o TensorRT (ms)   & -   & -    & 247.62   & -  & -   \\ \hline
% Inference Time w/ TensorRT (ms)   & 113.35   & 175.30    & 47.78   & 217.29  & 186.67   \\ \hline
% Accuracy w/o TensorRT(\%)       &$97.56^*$   &$91.27^*$  &$93.54$  &$95.42^*$  &$94.16^*$
% \\ \hline
% Accuracy w/ TensorRT(\%)       &95.72   &89.36  &88.40  &94.29  &93.06
% \\ \hline
% \end{tabular}
% %
% \vspace{0.1in}
% %
% \\
% Note: ``-'' means the model is not deployable directly on Jetson Nano, so it was run on the desktop workstation, GTX 1080Ti and its accuracy without the optimization through TensorRT was obtained.
% \label{table:modelInferenceTimeAndAccuracy}
% \end{table*}

\begin{figure}[b!]
	\centering
	\includegraphics[width=.9\columnwidth]{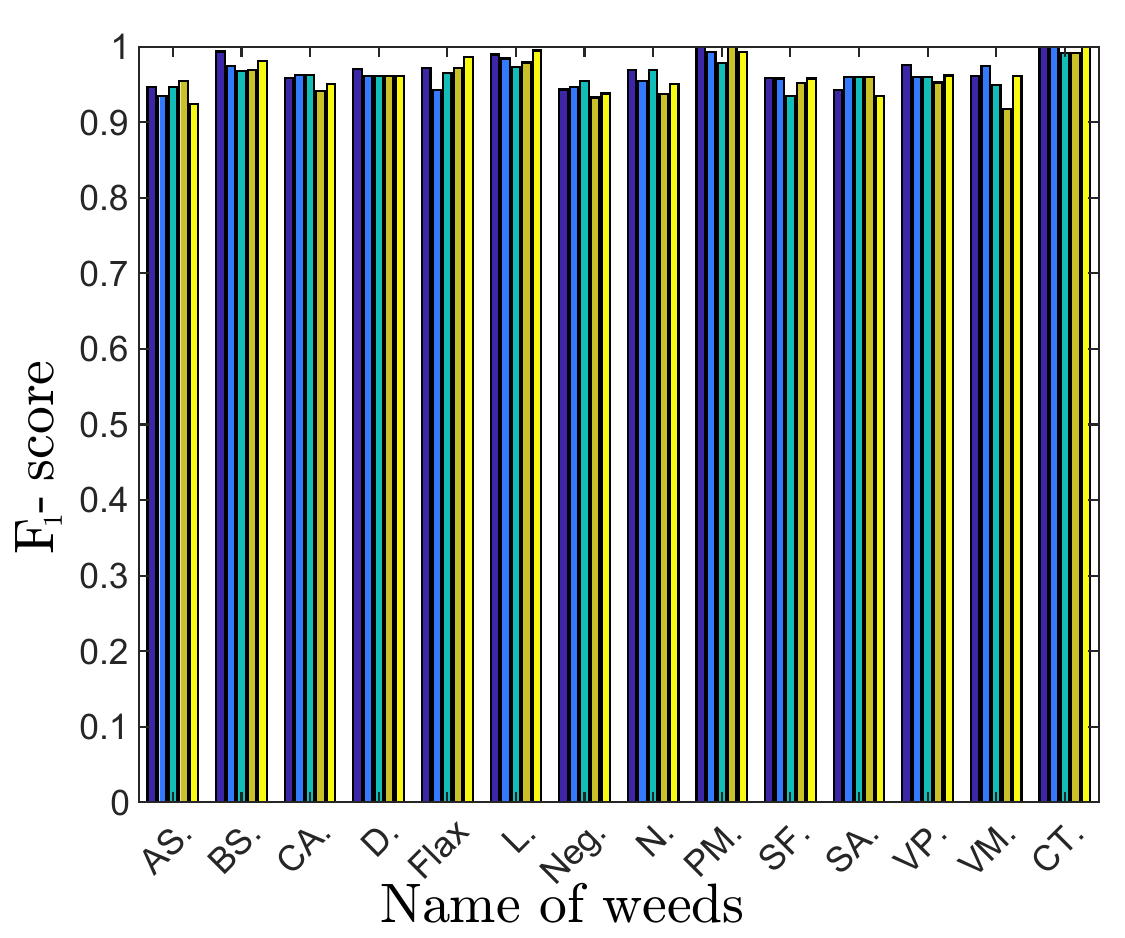}
	\caption{$\textit{F}_1$-score of 5 models per weed class (horizontal axis). The 5 models at each class from left to right are DenseNet121 (purple, 64 epochs), InceptionV3 (blue, 64 epochs), MobileNetV2 (Tiffany blue, 128 epochs), ResNet50 (dark yellow, 64 epochs), and Xception (bright yellow, 64 epochs), respectively. 
	}
	\label{fig:f1score}
\end{figure}
Fig. \ref{fig:f1score} displays the $F_1$-score of the five trained models on each kind of weed. All models perform well (above 90\% for all classes) considering the complexity of \textit{AIWeeds} as mentioned in Section \ref{sec:dataCollection}. The $F_1$-score of Neg. (backgrounds) class is relatively low. Looking carefully into the confusion matrix of all models, we find that most mispredictions are between Neg. and other weeds. This makes sense as the number of Neg. in \textit{AIWeeds} is double as other weeds, as shown in Table~\ref{table:weedAndNumber}. 
% more prone to being mixed with   and predicted wrongly. 
% This can also be seen from the confusion matrix in Fig.\ref{fig:confusion}
The targeted plant takes a smaller area out of the overall image. When we took pictures of it, we did not avoid other native non-targeted plants in view, so it would be more difficult to be identified from a noisy background.
% Additionally, the other typical mixture is between SA. and CA. as their leaf shapes and even 
% petioles are extremely similar at their early growth stages. 
One surprising finding is that BS. is narrow-leaf and looks similar to SF. from human recognition, especially at an early stage and in a noisy background, but were distinguished well by CNN models. This verifies the ability of CNNs to use a deep hierarchical structure to extract features.
% Also, interestingly, the number of input images does not always guarantee performance improvement. 
% For instance, Cirs. is not the one classified by models most accurately, even though its number of samples in \textit{AIWeeds} is 1.5 times that of other weed species. 
% We suspect that it could be due to large variation of light and background disturbance in our dataset and its resemblance to Sonch. at the early canopy stages. 
% Another point of note is MobileNetV2 realized excellent performance even compared with other more complicated models.

% \begin{figure}
% 	\centering
% 	\includegraphics[width=0.8\columnwidth]{fig6_confusion.pdf}
% 	\caption{ Confusion matrix of MobileNetV2 trained for 128 epochs before TensorRT-optimization. That after TensorRT-optimization is almost the same. All the numbers in this figure are percentages.
% 	}
% 	\label{fig:confusion}
% \end{figure}
\begin{figure}[t]
	\centering
	\includegraphics[width=\columnwidth]{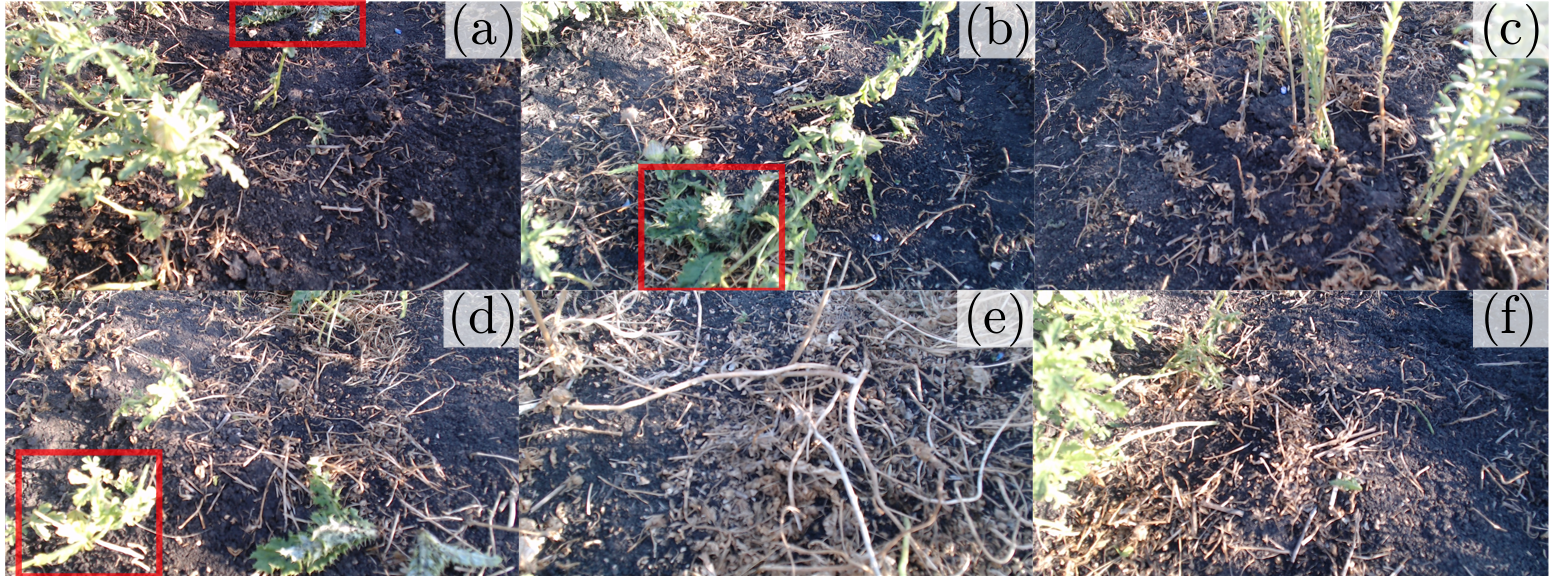}
	\caption{Images extracted from the video streaming on the robot. The prediction results of MobileNetV2 (128 epochs) model are (a) VM. (correct), (b) VM. (correct), (c) flax (correct), (d) CT. (correct), (e) Neg. (correct), and (f) CT. (wrong; ground truth is VM.).
	}
	\label{fig:fieldTest}
\end{figure}
\subsection{Flax Field Experiments} 
\label{subsec:robotexp}
% The inference time and accuracy of the above well-trained models before and after TensorRT-optimization (on GTX 1080Ti) as described in Section \ref{subsec:deployment} are displayed in Table \ref{table:modelInferenceTimeAndAccuracy}.
All of the above well-trained models besides MobileNetV2 cannot be deployed onto \textit{SAMBot} (with Jetson Nano). 
% It can be noticed that MobileNetV2 is more than five times faster with a 6\% drop in accuracy. 
% The accuracy change for MobileNetV2 is displayed in more detail in Fig. \ref{fig:confusion}. 
% The apparent loss of accuracy can be explained by the inherently simple structure of MobileNetV2, which tends to be optimized already before applying TensorRT. TensorRT optimization may potentially destroy some of MobileNetV2's internal structure initially designed for higher accuracy.
% Long latency is a fatal flaw in applications that require real-time video streaming. Relying on the shortest inference time, MobileNetV2 becomes the most competitive candidate for model deployment on \textit{SAMBot}. 
Considering the running speed of the robot, the resolution of the two cameras as shown in Figs.~\ref{fig:overview}(a1) and~\ref{fig:robot} is set to $384 \times 224$, the same as the input of CNN models. The frame rate is set to 10 frames per second. 
% The position of the camera on our robot is close to the ground, which helps avoid the overlapping problem in bird's eye view and thus improves the success rate of weed classification. 
Experimentally, with MobileNetV2 running, we ran \textit{SAMBot} in flax fields in North Dakota with medium weed density (with VM. and CT.) for 15m, and the robot was able to classify each weed at an average accuracy of 90\%. Meanwhile, it consumed less herbicide than commercial sprayers for the same spraying range (the flux rate of our customized pressurized sprayer is 78ml/min while 95.6ml/min for a commercial two-sprayer system). The status of weeds before and after herbicide spraying is shown in Figs.~\ref{fig:overview}(b1) and (b2), respectively. Fig.~\ref{fig:fieldTest} shows the field test results. The robot successfully recognized and sprayed VM. in (a) and (b) though there was an apparent disturbance from CT. (in rectangles). It also succeeded in detecting flax in (c), spraying CT. in (d) under VM.'s appearance. The 10\% failures mainly consist of cases where the leaves of VM. looked similar to CT. under strong light exposure as displayed in (f). 
% Autonomous navigation, recharging, and spraying realizations are not the focus of this work, so their details are omitted.  

\section{Conclusions}
\label{sec:conclusion}
In summary, we introduce the first large, realistic multiclass weed image dataset, \textit{AIWeeds}, with considerable variation (e.g. shadows, lighting and perspective change, different plant growth stages, and so on), collected entirely from \textit{in situ} in flax fields or gardens. It consists of flax, the 14 most common weeds, and backgrounds collected from North Dakota and California (U.S.) and middle China. Based on our dataset, we present baseline performance of five benchmark CNN models -- DenseNet121, InceptionV3, MobileNetV2, ResNet50, and Xception -- all of which perform well with an average $F_1$-score above 90\% on a highly variable dataset. Our low-cost, compact \textit{SAMBot} with a computational resource-limited onboard Jetson Nano is then run in flax fields with a medium density of Venus mallow and Canada thistle after the lightest MobileNetV2 being deployed. It realized a weed classification accuracy of 90\% with an inference time of 47.78ms. 
% Some failures might also be caused by the blurring imperfection of our camera. 
Evaluations on both \textit{AIWeeds} dataset and experiments in real fields demonstrate that our system is (i) applicable to under-canopy weeding, (ii) adaptable to unseen scenes, and (iii) able to robustly classify weeds at varying growth stages and  environments. Our work (including a dataset, multi-class weed classification pipeline, and experimental results) is a milestone towards under-canopy weed control in fields. The data and tools introduced in this paper enable various commercial mobile robots to detect, classify, and manage weeds of multiple types and thus reduce the amount of herbicide use by at least an order of magnitude.

% In the future, to make the real-time weed classification on the robot more reliable, we must solve the issue of motion blur. 
% Also, we will continue to contribute to the dataset collection by adding multi-labeled weed images and implement object-detection based methods for spraying multiple weeds in the robot\rq{}s view simultaneously.

%\addtolength{\textheight}{-12cm}   % This command serves to balance the column lengths
% on the last page of the document manually. It shortens
% the textheight of the last page by a suitable amount.
% This command does not take effect until the next page
% so it should come on the page before the last. Make
% sure that you do not shorten the textheight too much.

%%%%%%%%%%%%%%%%%%%%%%%%%%%%%%%%%%%%%%%%%%%%%%%%%%%%%%%%%%%%%%%%%%%%%%%%%%%%%%%%

\section*{Acknowledgments}
We acknowledge support from the NSF (Award \# IIS - 1925360) and the National Institute of Food and Agriculture, USDA (Award \# 2021-67022-34200). Special thanks to Afrina Rahman and Prof. Mukhlesur Rahman from North Dakota State University for their field preparation and herbicide spraying instructions and Mr. Tongqing Du for helping us build \textit{AIWeeds}.

%\appendix
%\input{appendixes}
\bibliography{Mendeley}% Produces the bibliography via BibTeX.
\end{document}